\documentclass[journal]{IEEEtran}
\usepackage{amsmath,amsfonts}
\usepackage{algorithmic}
\usepackage{algorithm}
\usepackage{array}
\usepackage{textcomp}
\usepackage{stfloats}
\usepackage{url}
\usepackage{verbatim}
\usepackage{graphicx}
\usepackage{cite}

\usepackage{booktabs}
\usepackage{multirow}
\usepackage[table]{xcolor}
\usepackage{array}
\usepackage{subfloat}
\usepackage[caption=false,font=scriptsize,labelfont=rm,textfont=rm]{subfig}
\usepackage[hidelinks]{hyperref}
\usepackage{pdfpages}
\begin{document}

\title{\emph{ThinkBLOX}: 3D Indoor Scene Generation with Progressive Reasoning}

\author{Yuan Xiao, Can Wang, Xiangyu Kong, and Jing Liao
\thanks{Yuan Xiao and Jing Liao are with the Department
of Computer Science, City University of Hong Kong, Hong Kong (e-mail: yuan.xiao.cs@outlook.com; jingliao@cityu.edu.hk).

Can Wang is with the School of Computing and Data Science, University of Hong Kong (e-mail: cwang355-c@my.cityu.edu.hk).

Xiangyu Kong is with the School of Computer Science, Beijing Information Science and Technology University
(e-mail: xykong@bistu.edu.cn).

The corresponding author is Jing Liao.}
}

\maketitle

\begin{abstract}
While traditional graphics methods often synthesize 3D indoor scenes autoregressively or hierarchically, recent vision-language model (VLM)-based generators predominantly adopt a one-shot paradigm where the full layout is planned at once. This one-shot approach often requires global re-optimization or complete reconstruction during interactive editing (e.g., inserting or moving objects) and can lead to physically or semantically poorly organized arrangements. 
To address these challenges, we propose \emph{ThinkBLOX}, a VLM-based progressive reasoning framework that iteratively designs and refines 3D scenes. ThinkBLOX treats layout generation as a state-conditioned, step-by-step reasoning-and-action process. To power this, we construct the ThinkBLOX-Data-200K dataset, containing 224,757 procedural placement pairs annotated with multi-view scene context, explicit Chain-of-Thought (CoT) rationales, and structured JSON layouts. Through supervised fine-tuning (SFT) on this dataset, the VLM learns to bridge the reasoning–action gap under incremental updates. Furthermore, recognizing that scene synthesis is inherently a multi-solution task where SFT suffers from reward conflict, we introduce Tier-Decoupled GDPO. This reinforcement learning scheme organizes heterogeneous rewards into distinct tiers, stabilizing policy optimization across physical validity, semantic plausibility, and reasoning–action consistency. Extensive experiments show that \emph{ThinkBLOX} significantly outperforms recent one-shot and iterative baselines in physical plausibility, semantic alignment, and interactive editability. Additionally, we show that it supports diverse applications, including both global and local generation and rearrangement of 3D scenes.
\end{abstract}


\section{Introduction}
\begin{figure*}[!t]
	\centering
	\subfloat[One-shot Scene Generation]
	{\label{fig:sub1}\includegraphics[width=1.62in]{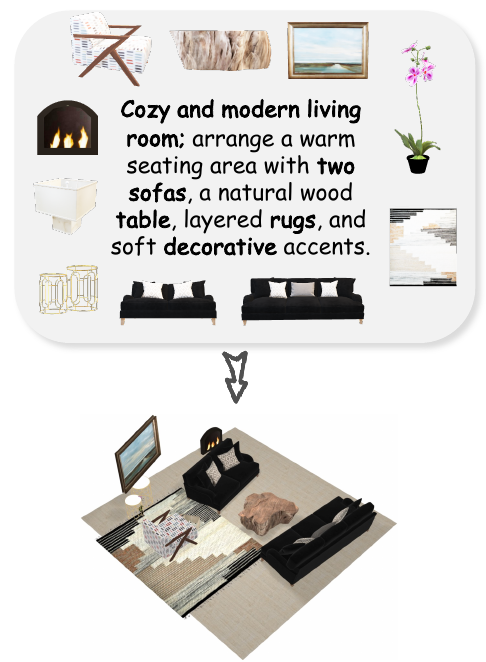}}\hspace{1.1in}
	\subfloat[Progressive Scene Generation]{\label{fig:sub2}\includegraphics[width=2.92in]{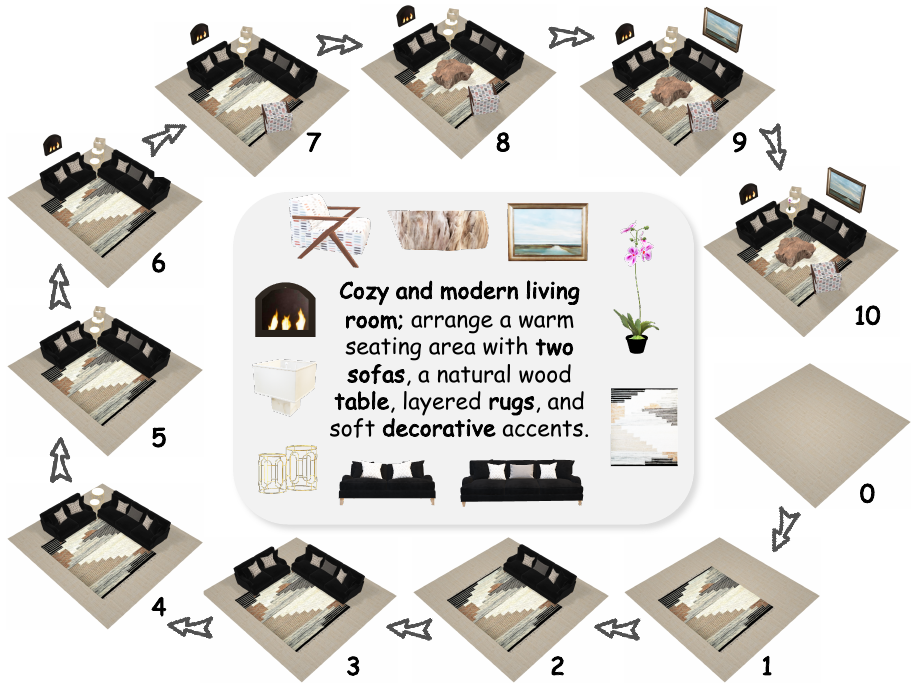}}
	\caption{Overall comparison between one-shot and progressive generation. (a) One-shot synthesis directly generates all object placements at once, which limits editability and typically requires global re-optimization for updates. (b) \emph{ThinkBLOX} follows a progressive process, placing objects sequentially to produce coherent intermediate layouts and better support interactive refinement.}
	\label{fig:multi}
\end{figure*}

3D indoor scene generation has emerged as a pivotal topic in computer vision and graphics, with applications spanning AR/VR, gaming, and embodied AI. In practical applications, generated layouts must not only be compatible with standard 3D engines for immediate deployment but also remain editable to accommodate evolving user requests. However, as shown in Figure~\ref{fig:multi}(a), most recent large-model-based generation methods \cite{feng2023layoutgpt,yang2024holodeck,pan2026metaspatial} are designed for \emph{one-shot} scene synthesis. Predicting an entire layout in a single pass limits flexible local editing and typically requires expensive global re-optimization or complete scene reconstruction for minor updates, restricting their practicality in interactive workflows.

Synthesizing scenes step-by-step is an intuitive strategy. In computer graphics, classical learning-based approaches have explored autoregressive placement \cite{paschalidou2021atiss,wang2021sceneformer} or hierarchical support trees \cite{fisher2012example,li2019grains}. While these methods generate layouts sequentially, they rely heavily on low-level geometric priors and struggle with open-vocabulary textual instructions or complex semantic reasoning. Conversely, while recent LLM/VLM methods excel at high-level semantic planning, they either force a one-shot layout output \cite{sun2025layoutvlm} or implement heuristic text-based loops \cite{wang2025chat2layout,yang2024llplace} that suffer from a pronounced \emph{reasoning–action gap}. Without synchronous visual state feedback and structured reasoning alignment, these text-only iterative models often output physically invalid positions or fail to anchor their textual rationales to exact 3D coordinates. This motivates the development of \emph{ThinkBLOX}, which introduces a progressive reasoning approach based on VLM, as illustrated in Figure~\ref{fig:multi}(b).

To enable VLMs to develop strong progressive reasoning abilities for layout generation, we first construct a dataset named ThinkBLOX-Data-200K, consisting of expert reasoning-action demonstrations organized as progressive placement pairs with multi-view scene context, Chain-of-Thought (CoT) rationales, and JSON layouts, for supervised fine-tuning (SFT) to train the VLM model to progressively infer object relationships and placements. While SFT provides a strong initialization, it is inherently limited for 3D scene generation and editing, which are naturally \emph{multi-solution} tasks. A single reference trajectory cannot cover the entire space of valid layouts. As a result, models trained solely with SFT may still generate physically invalid placements, semantically weak arrangements, or reasoning traces that are inconsistent with the actual object placements. 
Thus, we explore reinforcement learning (RL) and propose a Tier-Decoupled method to post-finetune the model. Specifically, we have designed multi-level reward functions covering physical validity, semantic plausibility, and reasoning-action consistency, and perform policy optimization with GDPO \cite{liu2026gdpo} under a novel tiered reward grouping strategy. We demonstrate that Tier-Decoupled GDPO organizes heterogeneous reward signals into structured tiers, reduces cross-objective interference, and enables stable and effective multi-objective optimization.
Extensive experiments demonstrate that \emph{ThinkBLOX} consistently achieves stronger physical plausibility, semantic alignment, and interactive editability than competitive LLM/VLM-based and constraint-based baselines, while substantially reducing the need for global re-layout and post-processing.
In summary, we make the following contributions:

\begin{itemize}
\item We propose \emph{ThinkBLOX}, a VLM-based progressive reasoning framework that enables the iterative design of 3D indoor scenes. This approach addresses the editing and quality limitations of one-shot scene generation methods, while supporting both global and local scene generation and rearrangement.

\item We construct the ThinkBLOX-Data-200K, a large-scale dataset consisting of 224,757 procedural placement pairs with multi-view scene context, CoT rationales, and structured JSON layouts, serving as a benchmark for progressive reasoning.

\item We introduce Tier-Decoupled GDPO, a novel RL approach that organizes heterogeneous rewards into distinct tiers for stable multi-objective policy optimization. It optimizes physical validity, semantic plausibility, and reasoning–action consistency, addressing the limitations of SFT in handling the multi-solution nature of 3D scene generation.
\end{itemize}

\section{Related Work}
\subsection{3D Indoor Scene Generation}
Early approaches to 3D indoor scene synthesis predominantly rely on geometric optimization under spatial constraints \cite{chang2014learning,chang2015text,feng2025casagpt} or data-driven generative distributions \cite{sun2024forest2seq}. To capture object dependencies, one prominent line of traditional work focuses on autoregressive generation, where objects are placed sequentially conditioned on previously generated entities. For instance, Deep Convolutional Priors \cite{wang2019deep} and PlanIT \cite{wang2019planit} utilize relation graphs and spatial networks to guide iterative placement, while SceneFormer \cite{wang2021sceneformer} and ATISS \cite{paschalidou2021atiss} leverage Transformer architectures to model the sequential distribution of object categories and bounding boxes. Another line explores hierarchical placement, organizing scenes via support relations or structural motifs, such as GRAINS \cite{li2019grains}, HLG \cite{wang2025hlg}, and HSM \cite{pun2026hsm}. Although these classical methods generate layouts progressively, they are strictly restricted to closed-set categories and lack the high-level semantic common sense required to interpret free-form user intents or open-vocabulary descriptions.

Recently, Large Language Models (LLMs) and Vision-Language Models (VLMs) have fundamentally advanced the controllability and semantic plausibility of scene layout generation \cite{feng2023layoutgpt,yang2024holodeck,kumaran2023scenecraft,ocal2024sceneteller,zhou2402gala3d,wang2023robogen,zhou2025scenex}. However, despite their strong semantic planning abilities, most representative LLM/VLM-based systems \cite{sun2025layoutvlm,pan2026metaspatial} predominantly pivot back to a one-shot paradigm. Their primary objective is to predict a complete layout in a single textual forward pass, often followed by post-hoc optimization or constraint handling. Consequently, they are poorly suited for interactive configurations where users wish to insert, move, or modify only a small subset of objects; in such cases, local edits inadvertently trigger global re-layout or full scene reconstruction. In contrast, \emph{ThinkBLOX} bridges the gap between these two paradigms, formulating scene synthesis as a progressive, vision-conditioned reasoning process that natively inherits the open-world flexibility of foundation models while supporting fine-grained interactive editing.

\subsection{Vision-Language Models for 3D Reasoning}
Recent progress in VLMs has improved spatial reasoning for visual and embodied tasks, making them a promising basis for layout planning. Existing work mainly improves spatial reasoning through (i) stronger spatial perception (e.g., 3D visual encoders or spatially supervised VLM adaptation) \cite{chen2024spatialvlm, cheng2024spatialrgpt}, and (ii) explicit reasoning augmentation, such as CoT reasoning, structured intermediate representations, symbolic discretization, and stepwise verification \cite{yang2023set,wang2025chat2layout}. While effective for understanding and visual reasoning, these methods do not directly address interactive 3D indoor layout generation, where reasoning must be translated into physically executable placements under incremental scene updates. \emph{ThinkBLOX} addresses this gap by combining state-aware VLM spatial reasoning, explicit CoT-to-action anchoring, and reward-driven policy optimization for interactive incremental 3D layout planning.
\subsection{RL for Enhancing Reasoning in Foundation Models}
RL has recently regained attention as a promising strategy for improving the reasoning capabilities of large foundation models \cite{guo2025deepseek}. Unlike SFT, which relies on fixed target outputs, RL optimizes model behavior from evaluative feedback and is therefore well suited to tasks with structural ambiguity or multiple valid solutions \cite{mondillo2025comparative}. In language models, RL from human feedback has been widely adopted to align model outputs with human preferences \cite{bai2022training}, and has played a central role in instruction-following systems \cite{wu2023brief}. Beyond human feedback, recent rule-based \cite{guo2025deepseek} and verifier-driven \cite{guo2025deepseek} reinforcement fine-tuning has shown strong results on code generation \cite{liu2025code}, and multi-step reasoning \cite{wang2025ragen}. In contrast, RL remains underexplored in multimodal spatial reasoning and 3D layout generation. A key challenge is that 3D layout generation is inherently \emph{multi-solution}: for a given instruction, multiple layouts may satisfy semantic intent and physical constraints. This makes pure SFT insufficient, as a single reference trajectory cannot capture the full solution space or the trade-offs among competing objectives (e.g., semantics, collisions, support, and layout compactness). To address this, \emph{ThinkBLOX} adopts a rule-driven RL formulation for VLM-based 3D layout generation and introduces Tier-Decoupled GDPO, a policy optimization scheme that stabilizes multi-objective learning via intra-group coupling and inter-tier decoupling. This design preserves semantic--physical dependencies while mitigating reward competition, enabling more robust and adaptive spatial decision making.

\begin{figure*}[!t]
  \centering
  \includegraphics[width=0.96\textwidth]{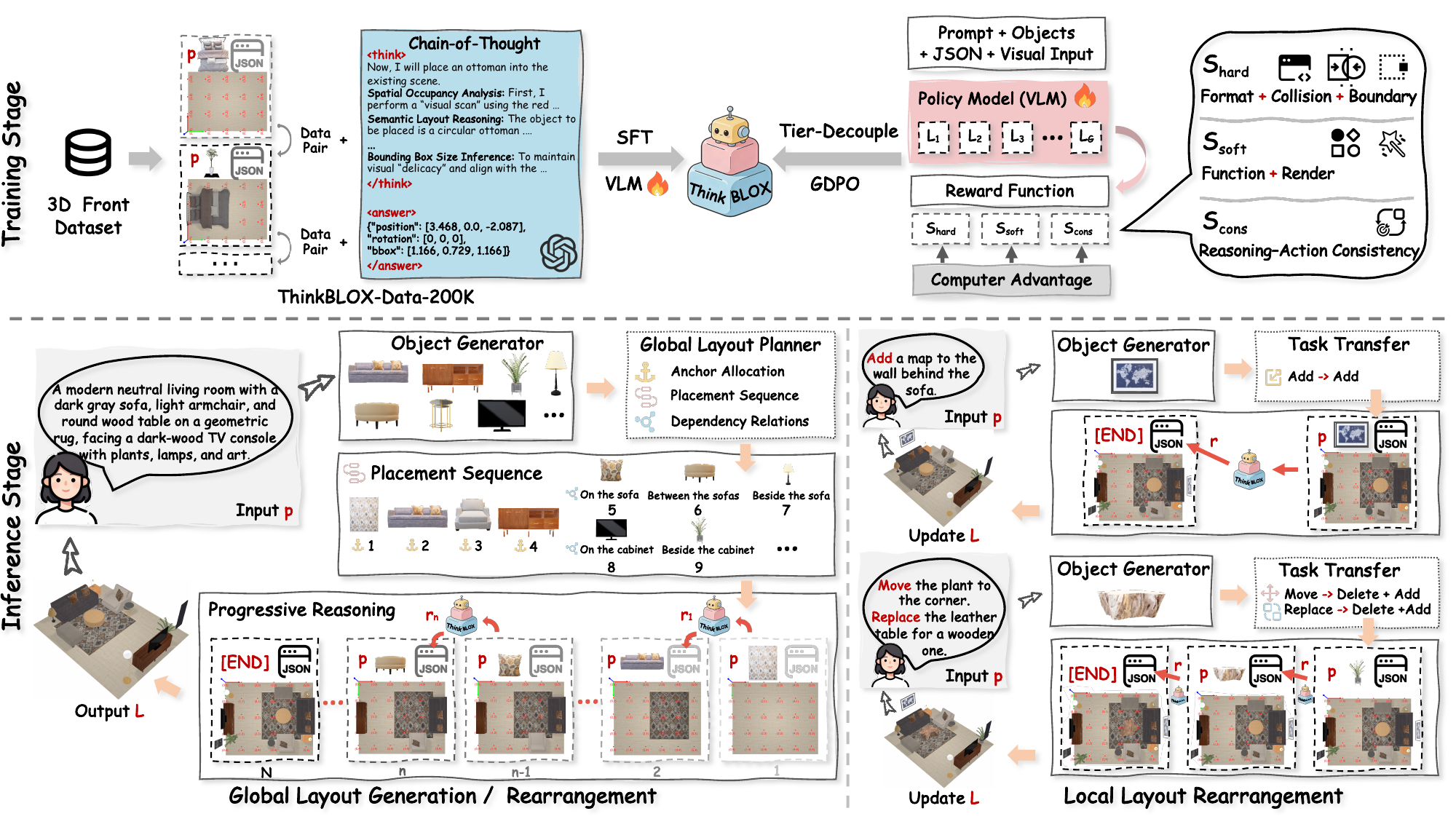}
  \caption{Overview of \emph{ThinkBLOX} framework. \textbf{Training:} We curate ThinkBLOX-Data-200K with multi-view context, CoT rationales, and JSON layouts, and train Qwen2.5-VL by SFT and Tier-Decoupled GDPO with tiered rewards ($S_{\text{hard}}$, $S_{\text{soft}}$, $S_{\text{cons}}$). \textbf{Inference:} Given a text prompt $\mathbf{p}$, \emph{ThinkBLOX} plans the placement order and progressively generates or edits layout $L$ through a CoT-guided reason-then-act process $\mathbf{r}_n$, supporting global generation/rearrangement and local rearrangement.}
  \label{fig:pipeline}
\end{figure*}

\section{Methodology}

Unlike previous works \cite{feng2023layoutgpt,yang2024holodeck,ccelen2024design,pan2026metaspatial,sun2025layoutvlm,ran2025direct} that treat 3D indoor scene generation as a one-shot process, we frame it as a progressive reasoning-action process.
Given a user input text prompt $\mathbf{p}$ defining the desired configuration of the room scene and the target object set $O=\{o_n\}_{n=1}^{N}$ inferred from the prompt, the model generates the final layout $L=\{(o_n,p_n,\theta_n,b_n)\}_{n=1}^{N}$ through progressive reasoning steps $\mathbf{r}_1, \mathbf{r}_2, \cdots, \mathbf{r}_N$, where $p_n \in \mathbb{R}^3$, $\theta_n$, and $b_n \in \mathbb{R}^3$ denote the position, orientation, and 3D bounding box of object $o_n$, respectively. Each intermediate reasoning step $\mathbf{r}_n$ is conditioned on the current scene context and is expressed as $\mathbf{r}_n \sim p(\mathbf{r}_n \mid \mathbf{p},L_{n-1},I_{n-1},o_n)$, where $L_{n-1}$ is the layout at step $n-1$, and $I_{n-1}$ is the visual input at step $n-1$, which comprises the rendered image of the current layout and the image of the target object to be placed. The placement of each object is extracted from the reasoning output $\mathbf{r}_n$ through the process $(p_n,\theta_n,b_n)=\mathrm{Parse}(\mathbf{r}_n)$. Here, \texttt{Parse} refers to the process of extracting structured information from the reasoning step $\mathbf{r}_n$. 
The layout is then updated iteratively as $L_n = L_{n-1} \cup \{(o_n,p_n,\theta_n,b_n)\}$. This process sequentially places the target objects into the scene, determining each object's position, orientation, and size based on the reasoning step, until the final layout is completed.

Figure~\ref{fig:pipeline} illustrates the framework of \emph{ThinkBLOX}.
In the training phase, \emph{ThinkBLOX} begins with SFT using stepwise reasoning-action demonstrations from the ThinkBLOX-Data-200K dataset (see \textbf{Sec.~\ref{sec:dataset}}), enabling the model to acquire progressive spatial reasoning skills. \emph{ThinkBLOX} is then augmented with the proposed Tier-Decoupled GDPO, extending its capabilities in spatial reasoning.
In the inference stage, \emph{ThinkBLOX} first plans a functional placement order for the target objects based on the available scene space. It then incrementally generates CoT reasoning trajectory $\mathbf{r}_n$ to guide the placement of each object $(p_n, \theta_n, b_n)$, conditioned on the user input, room context, and current partial layout. \emph{ThinkBLOX} supports a wide range of applications, including global layout generation/rearrangement and local layout rearrangement, while providing a more controllable, physically grounded, and reasoning-aligned approach compared to previous one-shot layout generation methods.

\subsection{Training with Progressive Reasoning}

\emph{ThinkBLOX} is trained in two progressive reasoning stages: SFT with progressive CoT reasoning, followed by post-finetuning of each reasoning step using Tier-Decoupled GDPO.

\subsubsection{SFT with Progressive CoT Reasoning}

We adopt Qwen2.5-VL-7B \cite{bai2025qwen25vltechnicalreport} as the backbone model and perform SFT on our progressive reasoning–action dataset to activate the model’s spatial planning capability.
At each placement step $n$, the model takes as input the user text prompt $\mathbf{p}$, the visual prompt $I_{n-1}$, the previous partial layout $L_{n-1}$ represented as a JSON file, and the current target object $o_n$, also specified in JSON format.
Instead of directly predicting geometric parameters $(p_n,\theta_n,b_n)$, the model is trained to generate CoT reasoning trajectory $\mathbf{r}_n$, from which the object placement parameters are derived via the previously introduced \texttt{Parse} operation. The training objective is the token-level cross-entropy loss.
This SFT formulation explicitly encourages the model to follow a reason-then-act paradigm for 3D object placement under an evolving scene context.

To facilitate understanding, we illustrate an example $r_n$ from the progessive training dataset ThinkBLOX-Data-200K in Figure~\ref{fig:pipeline}. 
$r_n$ corresponds to a single placement step and is formatted as a structured reasoning–action sequence consisting of a \texttt{<think>} segment followed by an \texttt{<answer>} segment. The CoT annotations are generated by GPT-4o \cite{wu2024gpt} based on ground-truth layouts under carefully designed prompts. More details about the data construction process are provided in \textbf{Sec.~\ref{sec:dataset}}. 

\subsubsection{Post-Finetuning with Tier-Decoupled GDPO}

Although our SFT provides a strong initialization, it remains insufficient for 3D scene generation, which are inherently multi-solution tasks where a single reference trajectory cannot cover the full space of valid layouts. To further improve robustness and encourage exploration of diverse valid solutions, we introduce a RL stage.
We design a three-tier reward structure consisting of Physical Validity, Semantic Aesthetics, and Cognitive Grounding.

\paragraph{Physical Validity}

This tier ensures the physical validity of the layout through three dedicated reward functions: 1) \textbf{Format Reward.}
To ensure that the model outputs remain parsable, we assign $R_{format}=0$ to valid outputs and $R_{format}=-1$ otherwise.
2) \textbf{Collision Penalty.}
To penalize geometric overlaps between the newly placed object and existing objects, we first define a collision score $C_{\mathrm{coll}} = \sum_{i=1}^{n} \rho_i$, where $\rho_i$ denotes the normalized overlap ratio between the new object and the $i$-th existing object. The collision reward is then defined as
$R_{coll} = -\tanh(\lambda C_{\mathrm{coll}})$, 
where $\lambda = 5.0$, $R_{coll}\in[-1,0]$, and $0$ indicates a collision-free placement.
3) \textbf{Boundary Adherence.}
To ensure that objects remain within the room boundary, we define a boundary penalty $R_{layout} = -\tanh(\text{distance})$, where $\text{distance}>0$ denotes the penetration depth into the exterior region. To avoid reward saturation when objects are placed far outside the valid area, we model the exterior space as an infinitely thick solid volume. As a result, $R_{layout}\in(-1,0]$ provides a continuous penalty for out-of-bound placements.

\paragraph{Semantic Aesthetics}

This tier evaluates the functional and semantic plausibility of object placement through two reward functions: 1) \textbf{Functional Alignment.} For objects with a strong functional orientation, we replace direct angle regression with vector alignment. Let $\mathbf{v}_{func}$ denote the unit functional vector of the object, i.e., the front-facing direction of the placed object after applying the predicted rotation. Let $p_{\text{cur}}$ and $p_{\text{tar}}$ denote the 3D positions of the current object and its associated target object, respectively, and let $\hat{p}=(x,y)$ denote projection onto the horizontal plane. We compute the target direction as $\mathbf{v}_{target}=\frac{\hat{p}_{\text{tar}}-\hat{p}_{\text{cur}}}{\|\hat{p}_{\text{tar}}-\hat{p}_{\text{cur}}\|_2}.$
The functional alignment reward is then defined as
$R_{align} = \mathbf{v}_{func} \cdot \mathbf{v}_{target}$,
which is equivalent to the cosine of the angle between the two normalized vectors. 2) \textbf{Semantic Consistency.}
To obtain a dense semantic reward, we query GPT-4o on whether the generated layout conforms to the instruction and use the log-probability of the ``Yes'' token:
$R_{render} = e^{\text{logprob}_{yes}}$, where $R_{render}\in[0,1]$ reflects the evaluator's confidence.

\paragraph{Cognitive Grounding}
We introduce a reasoning consistency reward to measure whether the generated reasoning trajectory $\mathbf{r}_n$ is consistent with the executed action $a_n$: $R_{\text{cons}}(\mathbf{r}_n, a_n) = - \mathrm{dist}(\mathrm{Parse}(\mathbf{r}_n), a_n)$, where \texttt{Parse} extracts planned numerical constraints from $\mathbf{r}_n$ via a template-guided parser with regular expression matching. Intuitively, if the reasoning specifies a placement plan (e.g., ``0.5m to the left of the bed''), but the executed action places the object elsewhere, the model receives a negative reward.

GDPO \cite{liu2026gdpo} enables stable training without requiring a value model and effectively handles multiple reward signals. We therefore adopt GDPO to optimize the policy. However, directly applying GDPO to our reward functions may still lead to unstable training, as our rewards are grouped into three tiers that optimize different aspects of the policy, while GDPO normalizes all rewards jointly. To address this, we perform policy optimization with GDPO under a tiered reward grouping scheme.
Let the three reward sets be $\{S_{hard}, S_{soft}, S_{cons}\}$, where $S_{hard}=\{R_{format}$, $R_{coll}, R_{layout}\}$,
$S_{soft}=\{R_{align}, R_{render}\}$,
and $S_{cons}=\{R_{cons}\}$.
These sets correspond to physical validity, semantic aesthetics, and cognitive grounding, respectively.
For each tier $k \in \{hard, soft, cons\}$, we first aggregate the rewards within that tier: $r_k^{(i,j)} = \sum_{r \in S_k} r^{(i,j)}$, where $(i,j)$ denotes the $j$-th sampled rollout for prompt $i$. This grouping preserves the internal structure of correlated rewards within the same tier. We then apply a group-relative normalization independently for each tier to obtain the tier-wise advantage:
\begin{equation}
A_k^{(i,j)}=
\frac{
r_k^{(i,j)}-\mathrm{mean}\{r_k^{(i,1)},\dots,r_k^{(i,G)}\}
}{
\mathrm{std}\{r_k^{(i,1)},\dots,r_k^{(i,G)}\}
},
\label{eq:intra-gdpo}
\end{equation}
where $G$ is the number of sampled rollouts in the group.
To combine different optimization objectives, we compute the final advantage as a weighted sum of tier-wise advantages:
\begin{equation}
A_{sum}^{(i,j)}=\sum_{k\in\{hard,soft,cons\}} w_k A_k^{(i,j)},
\label{eq:inter-gdpo}
\end{equation}
where $w_k$ controls the relative importance of each tier. Finally, we apply batch-wise normalization for training stability:
\begin{equation}
\hat{A}_{sum}^{(i,j)}=
\frac{
A_{sum}^{(i,j)}-
\mathrm{mean}\left\{
A_{sum}^{(i',j')} \mid i' \in D_{\text{Batch}},\ j'=1,\dots,G
\right\}
}{
\mathrm{std}\left\{
A_{sum}^{(i',j')} \mid i' \in D_{\text{Batch}},\ j'=1,\dots,G
\right\}
+\epsilon
},
\label{eq:inter-gdpo2}
\end{equation}
where $D_{\text{Batch}}$ denotes the current mini-batch, and $\epsilon$ is a small positive constant for stability. This design couples correlated rewards within each tier while decoupling different tiers during normalization. Consequently, hard physical constraints, soft semantic preferences, and reasoning–action consistency can be jointly optimized without collapsing into a single undifferentiated training signal.

\subsection{Inference with CoT}

In the inference stage as shown in Figure~\ref{fig:pipeline}, \emph{ThinkBLOX} generates layouts through a CoT-guided reason-then-act process. Given the user prompt $\mathbf{p}$, our method first extracts the target object sets $O$ from $\mathbf{p}$, represented as a JSON file that records the ID and description of each object. The model then performs a global layout planning step before progressive placement. In this step, it first identifies anchor objects, namely large and function-defining items such as beds, sofas, dining tables, and cabinets, which determine the main structure and functional regions of the room. These anchor objects are selected based on their semantic category and spatial importance in the user instruction. The model then determines a coarse placement order following a coarse-to-fine strategy: anchor objects are placed first, followed by functionally related furniture and supporting items, and finally smaller decorative objects. At the same time, the model establishes dependency relations among objects according to their functional associations and intended usage, for example, chairs with tables, lamps with bedside tables, and decorative items with supporting surfaces. This planning step provides a coherent structural scaffold for the subsequent stepwise reason-then-act placement process.

\emph{ThinkBLOX} then performs a progressive reasoning process. At step $n$, given the user prompt $\mathbf{p}$, the visual prompt of the scene $I_{n-1}$, the partial layout $L_{n-1}$, and the current target object $o_n$, \emph{ThinkBLOX} first reasons about the trajectory $\mathbf{r}_n$, and then parses the placement parameters of object $o_n$ as
$(p_n,\theta_n,b_n)$.
This reasoning step determines where the object can be placed, why the placement is functionally and semantically appropriate, and how the object should be oriented and scaled with respect to the surrounding scene.
The reasoning process is repeatedly grounded in the evolving scene state and directly translated into executable placement actions. Through this autoregressive CoT-guided inference process, \emph{ThinkBLOX} maintains physical plausibility, semantic coherence, and reasoning–action consistency throughout both scene generation from scratch and local scene editing.

\subsection{Support for Diverse Applications}

Our progressive reasoning paradigm naturally supports diverse interactive applications without requiring any redesign of the inference pipeline. We categorize these applications into two types: global layout generation/rearrangement and local layout generation/rearrangement.

For global layout generation/rearrangement, \emph{ThinkBLOX} operates on the entire scene and progressively constructs or reorganizes the layout according to user intent. In generation, the model starts from an empty or minimally initialized room and sequentially places objects following a function-aware planning order. In rearrangement, it updates the scene layout under new instructions while maintaining coherence with room structure and inter-object relationships.

For local layout generation/rearrangement, \emph{ThinkBLOX} focuses on a user-specified object or region and updates only the relevant part of the layout instead of re-optimizing the whole scene. This formulation supports several object-level operations: local re-placement, object addition, object movement, and object replacement. These operations adjust object pose, insert new objects, relocate existing ones, or replace the target object while maintaining compatibility with the surrounding environment.

\section{ThinkBLOX-Data-200K for Progressive Reasoning}
\label{sec:dataset}

\begin{figure*}[!t]
  \centering
  \includegraphics[width=0.96\textwidth]{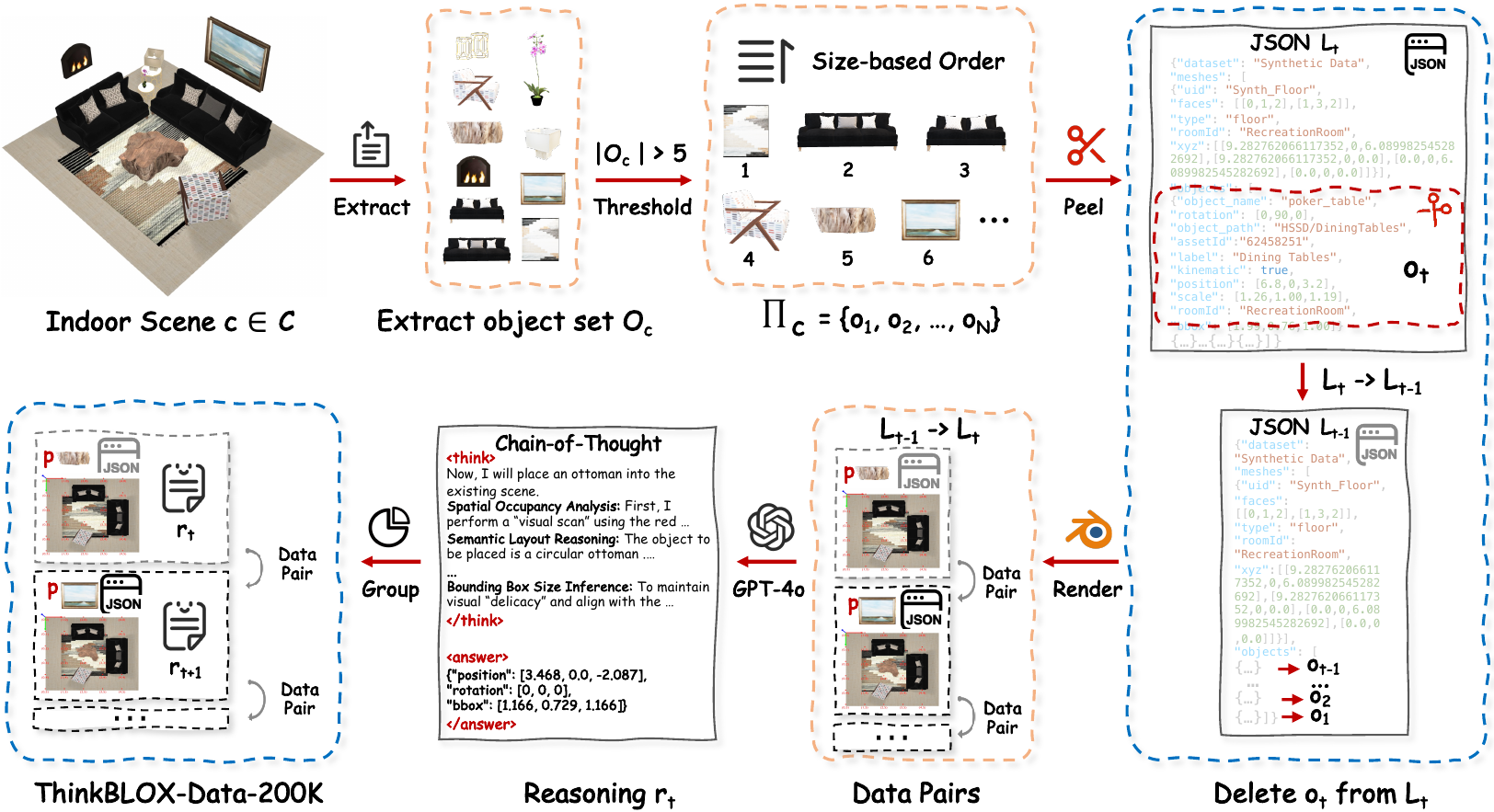}
  \caption{Construction Pipeline of ThinkBLOX-Data-200K. Starting from an indoor scene $c \in \mathcal{C}$, where $\mathcal{C}$ denotes the set of source scenes from IL3D, we first extract its object set $O_c$. Scenes with fewer than five objects are discarded. The remaining objects are then sorted by a size-based heuristic to obtain an ordered placement sequence $\Pi_c=\{o_1,o_2,\dots,o_N\}$, where $o_t$ denotes the $t-th$ object to be placed. For each step $t$, we remove the target object $o_t$ from the full layout $L_t$ to construct the previous partial layout $L_{t-1}$, render multimodal observations based on the pair $(L_{t-1}, o_t)$, and pair them with the textual prompt $\mathbf{p}$ describing the intended scene.
  GPT-4o is then used to generate the corresponding reasoning $\mathbf{r}_t$, including both chain-of-thought annotations and structured placement output. Finally, all valid pairs are grouped to form ThinkBLOX-Data-200K. Here, $L_t$ denotes the layout state containing objects up to step $t$, $L_{t-1}$ is the preceding partial scene, and $\mathbf{r}_t$ is the reasoning associated with placing $o_t$ into $L_{t-1}$.}
  \label{fig:data pipeline}
\end{figure*}

Existing layout generation benchmarks \cite{fu20213d,feng2023layoutgpt,lin2024instructscene,zhou2025il3d} often struggle with high-fidelity spatial reasoning due to their reliance on one-shot mappings rather than progressive reasoning. For example, IL3D \cite{zhou2025il3d} provides rich geometric information but does not support the progressive reasoning needed to synthesize complex and functionally coherent environments.
Thus, we construct ThinkBLOX-Data-200K, a large-scale dataset with 224,757 procedural placement pairs derived from IL3D. Each sample encodes how and why an object should be integrated into an evolving scene, represented by the current layout and target object, together with a multimodal rationale consisting of CoT reasoning and a structured JSON layout. Our curation pipeline includes four key stages. Figure~\ref{fig:data pipeline} provides an overview of the construction pipeline of ThinkBLOX-Data-200K.

\paragraph{Raw Data Collection}
We extract high-fidelity indoor scenes from IL3D \cite{zhou2025il3d}, which provides diverse furniture categories and precise spatial annotations. To encourage deeper spatial reasoning, we select complex scenes with at least five objects, ensuring rich context for functional and aesthetic layout planning.
\paragraph{Sequential Scene Decomposition}
Unlike traditional methods that treat a scene holistically, we decompose layouts into sequential object-level placement pairs. Following a large-to-small heuristic, structural anchors (e.g., beds, sofas) are placed first, followed by functional accessories (e.g., lamps, chairs), allowing the model to learn spatial dependencies through incremental construction.
\paragraph{CoT Reasoning Synthesis}
Leveraging GPT-4o, we generate high-quality CoT rationales for each placement. By analyzing scene context, including functional surfaces and directional constraints, the reasoning explains the position and orientation of each new object. This encourages layouts derived from structured spatial planning rather than simple coordinate regression.
\paragraph{Rule-based Quality Refinement}
We implement a multi-stage filtering process to ensure the reliability of the reasoning chains. First, we verify the structural integrity of the \texttt{<think>} and \texttt{<answer>} tags. Second, we filter out samples with fewer than three reasoning steps or insufficient word counts. Finally, a consistency check is performed; only high-quality samples, where the Levenshtein similarity between the original and re-predicted answers exceeds 0.8, are retained.

Please refer to the supplementary material for more dataset details.

\section{Experiments}

\subsection{Settings}

\subsubsection{Evaluation Metrics}
We assess the quality of synthesized 3D layouts through two primary lenses: physical plausibility and semantic consistency with the input instructions. To quantify physical plausibility, we employ the Collision-Free (CF) and In-Boundary (IB) scores. 
For semantic coherence, we use Positional (Pos.) and Rotational (Rot.) Coherency to determine how well the layout aligns with the textual prompt. Furthermore, following LayoutVLM \cite{sun2025layoutvlm}, we adopt the Physically-Grounded Semantic Alignment (PSA) score to simultaneously assess physical and semantic quality. The PSA score is computed by integrating the GPT-4o-based semantic rating, where GPT-4o serves as a human-aligned evaluator \cite{wu2024gpt}, with the physical plausibility metric. The final score ranges from 0 to 100, where 0 indicates a failure to achieve a feasible placement.
\subsubsection{Baselines} We evaluate our method against the following open‑source baselines: LLM‑based methods such as LayoutGPT \cite{feng2023layoutgpt}, Holodeck \cite{yang2024holodeck} and I‑Design \cite{ccelen2024design}, and VLM‑based methods such as MetaSpatial \cite{pan2026metaspatial}.

\subsubsection{Details of Implementation}
We initialize \emph{ThinkBLOX} from Qwen2.5-VL-7B \cite{bai2025qwen25vltechnicalreport}. In the SFT stage, the VLM is trained for 10K steps using LoRA $(rank=16, alpha=32)$ with a learning rate of $2 \times 10^{-5}$. In the Tier-Decoupled GDPO post-finetuning stage, the model is further optimized for 2K steps with a learning rate of $2 \times 10^{-5}$. We set the KL coefficient to $\beta = 0.004$ and the sample group size to 8 per prompt. All training is conducted on 8 NVIDIA RTX PRO 6000 GPUs, each with 96GB memory.

\subsection{Main Results}
\subsubsection{Quantitative Comparisons.}

\begin{table*}[!t]
\centering
\scriptsize
\setlength{\tabcolsep}{2.2pt}
\renewcommand{\arraystretch}{1.08}
\caption{Quantitative comparison on the layout generation benchmark across 11 room types. We report physical validity (CF$\uparrow$, IB$\uparrow$), semantic accuracy (Pos.$\uparrow$, Rot.$\uparrow$), and the overall PSA$\uparrow$ score. \emph{ThinkBLOX} achieves the best overall performance and generalizes consistently across diverse room categories.}
\label{tab:main_results}

\resizebox{\textwidth}{!}{
\begin{tabular}{l *{4}{cccc>{\columncolor{blue!5}}c}}
\toprule
 & \multicolumn{5}{c}{\textbf{Bedroom}}
 & \multicolumn{5}{c}{\textbf{Living Room}}
 & \multicolumn{5}{c}{\textbf{Dining Room}}
 & \multicolumn{5}{c}{\textbf{Bookstore}} \\
\cmidrule(lr){2-6}
\cmidrule(lr){7-11}
\cmidrule(lr){12-16}
\cmidrule(lr){17-21}
Methods
& CF$\uparrow$ & IB$\uparrow$ & Pos.$\uparrow$ & Rot.$\uparrow$ & PSA$\uparrow$
& CF$\uparrow$ & IB$\uparrow$ & Pos.$\uparrow$ & Rot.$\uparrow$ & PSA$\uparrow$
& CF$\uparrow$ & IB$\uparrow$ & Pos.$\uparrow$ & Rot.$\uparrow$ & PSA$\uparrow$
& CF$\uparrow$ & IB$\uparrow$ & Pos.$\uparrow$ & Rot.$\uparrow$ & PSA$\uparrow$ \\
\midrule
LayoutGPT
& 100.0 & 66.7 & 85.7 & 85.9 & 52.2
& 44.4 & 11.1 & 74.7 & 64.4 & 9.6
& 88.9 & 22.2 & 76.0 & 68.9 & 14.8
& 88.9 & 55.6 & 80.9 & 79.4 & 35.9 \\
Holodeck
& 88.9 & 22.2 & 69.3 & 67.9 & 14.1
& 77.8 & 0.0 & 66.3 & 55.6 & 0.0
& 88.9 & 0.0 & 38.0 & 36.6 & 0.0
& 55.6 & 0.0 & 65.7 & 59.0 & 0.0 \\
I-Design
& 100.0 & 77.8 & 72.1 & 65.4 & 51.5
& 33.3 & 11.1 & 62.6 & 46.7 & 0.0
& 88.9 & 66.7 & 76.4 & 66.4 & 34.8
& 66.7 & 11.1 & 68.1 & 69.4 & 5.2 \\
MetaSpatial
& 88.9 & 77.8 & 82.6 & 55.8 & 53.3
& 22.2 & 55.6 & 63.3 & 42.7 & 9.1
& 100.0 & 88.9 & 39.2 & 47.4 & 37.9
& 55.6 & 88.9 & 73.6 & 67.5 & 34.7 \\
\midrule
\textbf{\emph{ThinkBLOX}}
& 88.9 & 88.9 & 77.9 & 76.4 & \textbf{56.7}
& 22.2 & 66.7 & 65.8 & 49.5 & \textbf{9.9}
& 100.0 & 100.0 & 57.9 & 53.4 & \textbf{56.6}
& 66.7 & 88.9 & 79.1 & 77.3 & \textbf{42.1} \\
\bottomrule
\end{tabular}
}

\vspace{0.8em}

\resizebox{\textwidth}{!}{
\begin{tabular}{l *{4}{cccc>{\columncolor{blue!5}}c}}
\toprule
 & \multicolumn{5}{c}{\textbf{Buffet Restaurant}}
 & \multicolumn{5}{c}{\textbf{Children Room}}
 & \multicolumn{5}{c}{\textbf{Classroom}}
 & \multicolumn{5}{c}{\textbf{Computer Room}} \\
\cmidrule(lr){2-6}
\cmidrule(lr){7-11}
\cmidrule(lr){12-16}
\cmidrule(lr){17-21}
Methods
& CF$\uparrow$ & IB$\uparrow$ & Pos.$\uparrow$ & Rot.$\uparrow$ & PSA$\uparrow$
& CF$\uparrow$ & IB$\uparrow$ & Pos.$\uparrow$ & Rot.$\uparrow$ & PSA$\uparrow$
& CF$\uparrow$ & IB$\uparrow$ & Pos.$\uparrow$ & Rot.$\uparrow$ & PSA$\uparrow$
& CF$\uparrow$ & IB$\uparrow$ & Pos.$\uparrow$ & Rot.$\uparrow$ & PSA$\uparrow$ \\
\midrule
LayoutGPT
& 100.0 & 33.3 & 81.2 & 83.3 & 26.9
& 100.0 & 0.0 & 80.9 & 82.6 & 0.0
& 88.9 & 0.0 & 76.3 & 66.7 & 0.0
& 100.0 & 22.2 & 87.8 & 85.2 & 17.8 \\
Holodeck
& 77.8 & 11.1 & 47.7 & 42.4 & 7.4
& 77.8 & 22.2 & 72.7 & 70.0 & 18.7
& 33.3 & 0.0 & 45.2 & 38.6 & 0.0
& 100.0 & 0.0 & 66.1 & 59.7 & 0.0 \\
I-Design
& 100.0 & 55.6 & 63.5 & 57.1 & 35.2
& 77.8 & 55.6 & 78.1 & 75.1 & 34.8
& 55.6 & 11.1 & 50.7 & 47.0 & 0.0
& 88.9 & 22.2 & 74.0 & 70.7 & 8.9 \\
MetaSpatial
& 77.8 & 55.6 & 53.2 & 51.3 & 37.5
& 100.0 & 88.9 & 73.6 & 73.8 & 67.7
& 77.8 & 55.6 & 52.4 & 49.7 & 37.4
& 88.9 & 55.6 & 68.7 & 63.4 & 37.5 \\
\midrule
\textbf{\emph{ThinkBLOX}}
& 100.0 & 77.8 & 69.4 & 59.4 & \textbf{48.9}
& 100.0 & 100.0 & 83.2 & 72.9 & \textbf{84.7}
& 66.7 & 100.0 & 66.1 & 53.8 & \textbf{45.6}
& 88.9 & 88.9 & 74.6 & 81.9 & \textbf{57.3} \\
\bottomrule
\end{tabular}
}

\vspace{0.8em}

\resizebox{\textwidth}{!}{
\begin{tabular}{l *{4}{cccc>{\columncolor{blue!5}}c}}
\toprule
 & \multicolumn{5}{c}{\textbf{Deli}}
 & \multicolumn{5}{c}{\textbf{Florist Shop}}
 & \multicolumn{5}{c}{\textbf{Game Room}}
 & \multicolumn{5}{c}{\textbf{Average}} \\
\cmidrule(lr){2-6}
\cmidrule(lr){7-11}
\cmidrule(lr){12-16}
\cmidrule(lr){17-21}
Methods
& CF$\uparrow$ & IB$\uparrow$ & Pos.$\uparrow$ & Rot.$\uparrow$ & PSA$\uparrow$
& CF$\uparrow$ & IB$\uparrow$ & Pos.$\uparrow$ & Rot.$\uparrow$ & PSA$\uparrow$
& CF$\uparrow$ & IB$\uparrow$ & Pos.$\uparrow$ & Rot.$\uparrow$ & PSA$\uparrow$
& CF$\uparrow$ & IB$\uparrow$ & Pos.$\uparrow$ & Rot.$\uparrow$ & PSA$\uparrow$ \\
\midrule
LayoutGPT
& 88.9 & 0.0 & 77.2 & 77.9 & 0.0
& 66.7 & 33.3 & 81.6 & 80.2 & 18.3
& 55.6 & 22.2 & 87.0 & 82.9 & 6.7
& 83.8 & 24.2 & 80.8 & 78.0 & 16.6 \\
Holodeck
& 88.9 & 33.3 & 73.9 & 63.7 & 24.4
& 66.7 & 0.0 & 73.2 & 64.7 & 0.0
& 55.6 & 22.2 & 60.7 & 58.0 & 0.0
& 77.8 & 8.1 & 62.8 & 55.6 & 5.6 \\
I-Design
& 88.9 & 22.2 & 67.8 & 65.9 & 10.4
& 77.8 & 0.0 & 75.5 & 68.3 & 0.0
& 66.7 & 44.4 & 62.8 & 58.9 & 17.0
& 76.8 & 34.3 & 68.3 & 62.8 & 18.0 \\
MetaSpatial
& 77.8 & 55.6 & 78.2 & 72.4 & 39.8
& 55.6 & 88.9 & 76.7 & 65.3 & 38.5
& 66.7 & 55.6 & 67.4 & 66.8 & 31.4
& 73.8 & 69.7 & 66.3 & 59.6 & 38.6 \\
\midrule
\textbf{\emph{ThinkBLOX}}
& 88.9 & 88.9 & 85.2 & 79.5 & \textbf{61.4}
& 77.8 & 100.0 & 78.6 & 72.8 & \textbf{56.7}
& 77.8 & 88.9 & 68.3 & 72.2 & \textbf{49.6}
& 79.8 & 89.9 & 73.3 & 68.1 & \textbf{51.8} \\
\bottomrule
\end{tabular}
}

\end{table*}

\begin{figure*}[!t]
  \centering
  \includegraphics[width=0.96\textwidth]{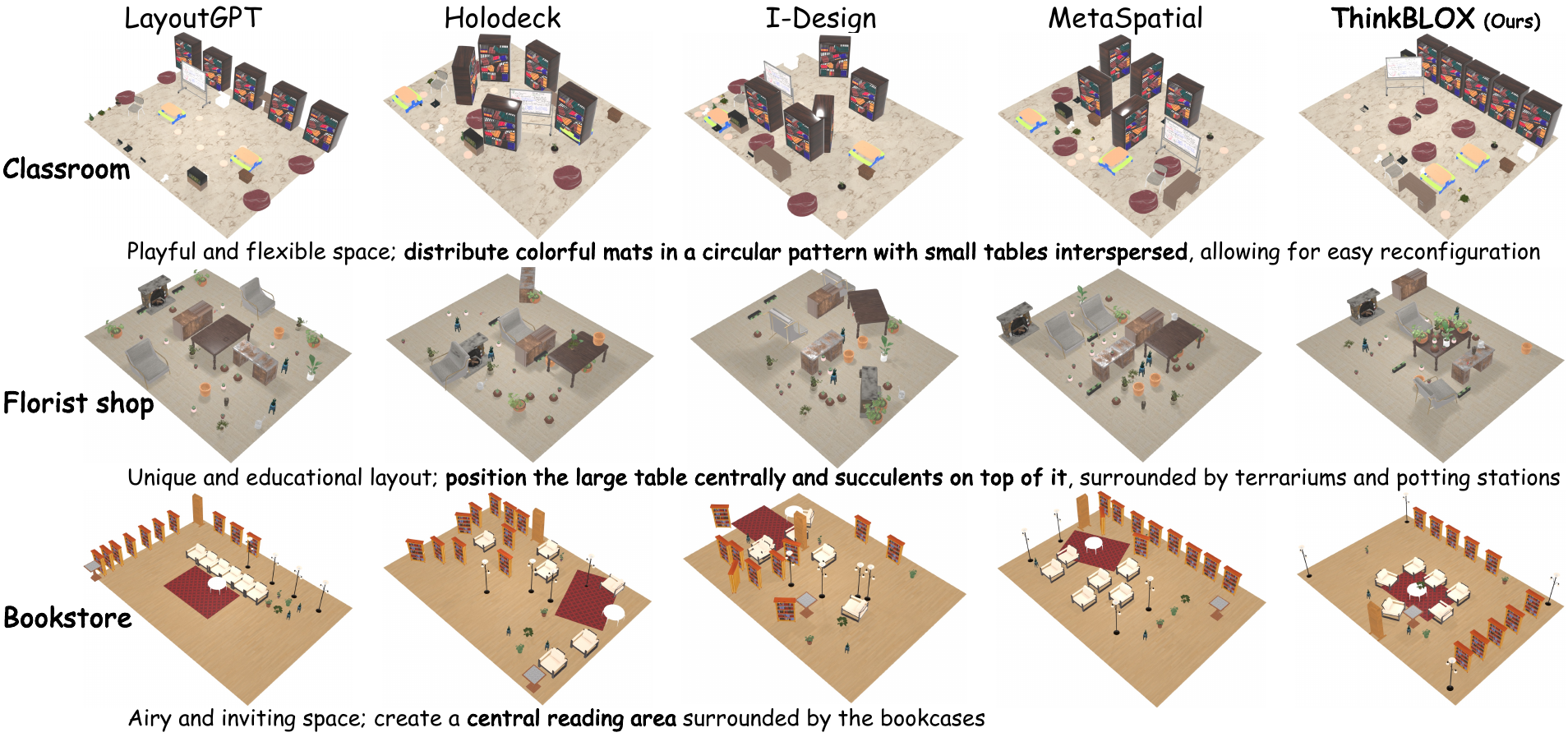}
  \caption{Qualitative comparison on the benchmark. We compare with baseline methods in generating layouts based on detailed prompts. Our method is able to generate layouts that closely follow the instructions and adhere to physical constraints.}
  \label{fig:Qualitative-Comparison}
\end{figure*}

\begin{figure*}[!t]
  \centering
  \includegraphics[width=0.96\textwidth]{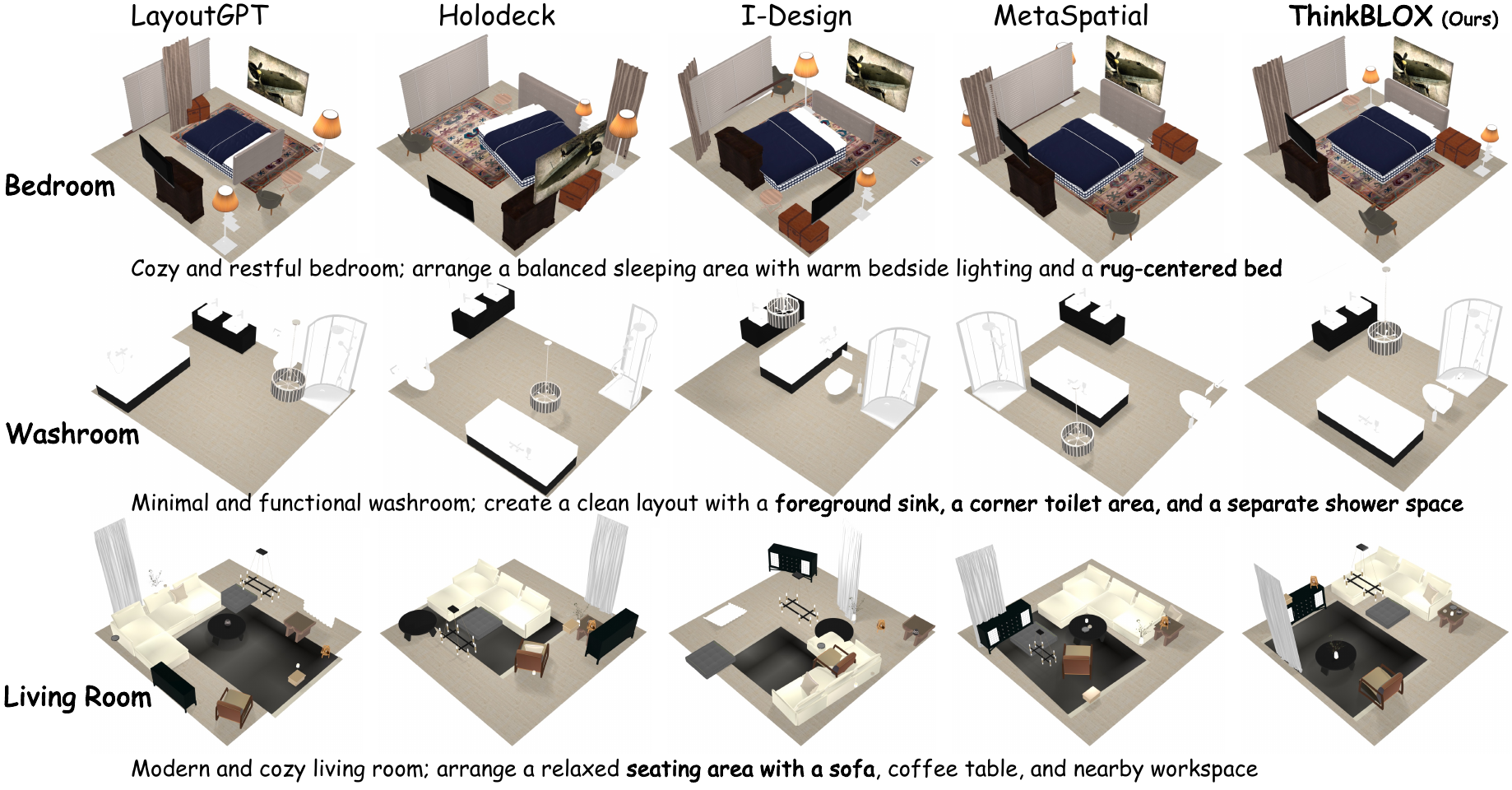}
  \caption{Qualitative comparison on ThinkBLOX-Data-200K. We report representative results on test dataset across diverse room types. Compared to baselines, \emph{ThinkBLOX} better follows detailed prompts while maintaining physically plausible layouts.}
  \label{fig:Qualitative-Comparison-ourdata}
\end{figure*}

Our method achieves the best overall PSA score, as shown in Table~\ref{tab:main_results}. This gain stems from our progressive generation paradigm, which makes step-wise, context-conditioned decisions and thus yields layouts that are both physically valid and semantically coherent. Compared to one-shot baselines, our approach achieves a better balance between physical plausibility and semantic coherence, leading to a clear advantage on the comprehensive PSA measure. In particular, LayoutGPT and I-Design often trade physical feasibility for semantic alignment, while Holodeck degrades under many assets and rigid constraints. Overall, our method reduces boundary violations while preserving positional and rotational coherence.

\subsubsection{Qualitative Comparisons.}

In Figure~\ref{fig:Qualitative-Comparison}, we present qualitative examples on the benchmark \cite{sun2025layoutvlm} generated by each method, and in Figure~\ref{fig:Qualitative-Comparison-ourdata}, we show qualitative results on our ThinkBLOX-Data-200K test set.
Our method performs particularly well in dense indoor scenes, benefiting from the proposed progressive reasoning process. For instance, in the florist shop, \emph{ThinkBLOX} can follow the instruction to place over 13 plants on tables. In contrast, one-shot baselines often exhibit an imbalanced trade-off between semantic alignment and physical feasibility, leading to issues such as collisions or boundary violations. More qualitative comparison results on both the benchmark and ThinkBLOX-Data-200K are provided in the supplementary material.

\begin{figure*}[!t]
  \centering
  \includegraphics[width=0.96\textwidth]{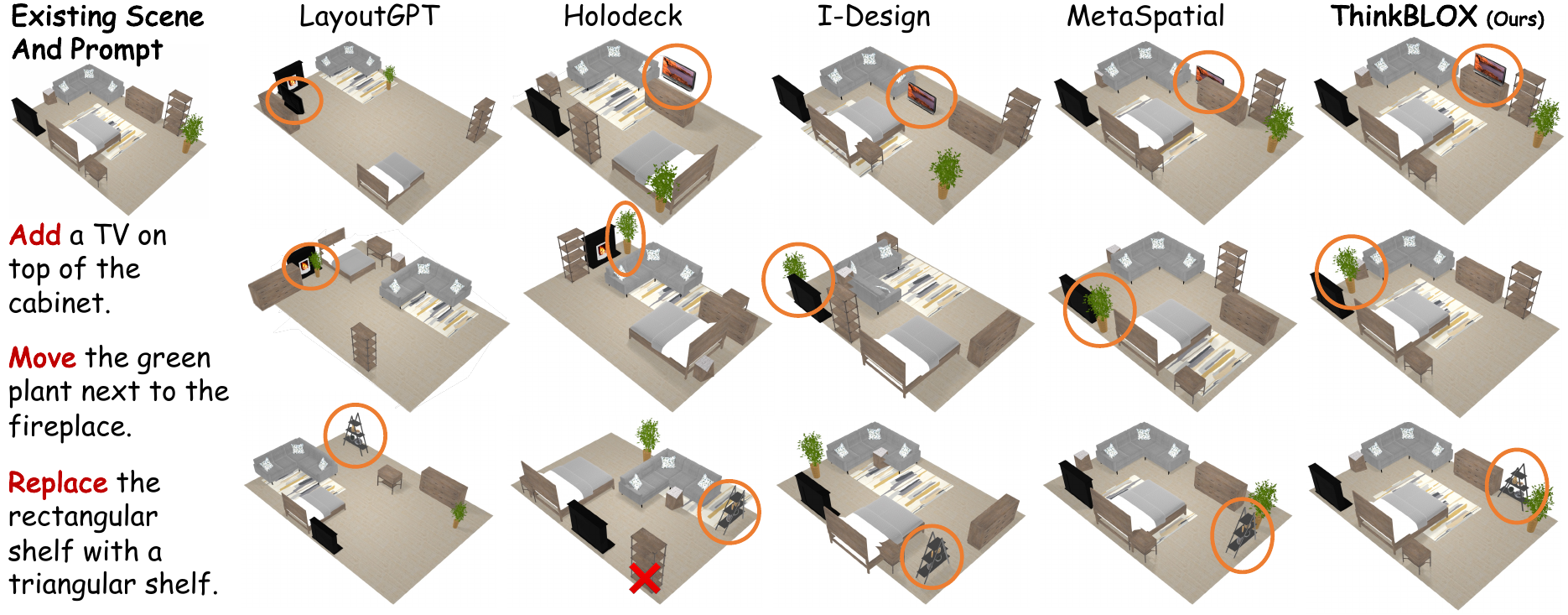}
  \caption{Local layout rearrangement comparison. Given an existing scene and an editing prompt, we compare add/move/replace edits across methods (orange circles). Red crosses indicate cases where a method replaces the wrong object. \emph{ThinkBLOX} follows instructions while preserving the remaining layout, without global re-layout.
  }
  \label{fig:edit}
\end{figure*}

\begin{table}[!t]
\centering
\caption{Cross-model evaluation with independent judges.}
\label{tab:cross_model_eval}
\resizebox{\linewidth}{!}{
\begin{tabular}{lcccc>{\columncolor{blue!5}}c}
\toprule
Method & LayoutGPT & Holodeck & I-Design & MetaSpatial & \emph{ThinkBLOX}  \\
\midrule
Gemini 3 & 18.4 & 6.2 & 19.8 & 39.4 & \textbf{52.5} \\
Qwen3-VL-30B & 14.9 & 7.4 & 12.6 & 26.7 & \textbf{50.8} \\
\bottomrule
\end{tabular}}
\end{table}

\subsubsection{Evaluator Robustness.}
Since automatic semantic evaluation may be sensitive to the choice of evaluator, we further validate the comparison using two independent VLM judges, Gemini 3 \cite{gemini3} and Qwen3-VL-30B \cite{qwen3vl}. As shown in Table~\ref{tab:cross_model_eval}, \emph{ThinkBLOX} consistently achieves the highest scores under both evaluators. This suggests that the observed gains are robust to the choice of evaluator and are not tied to a specific GPT-4o-based preference.

\subsection{Layout Editing.}

We further demonstrate the capability of \emph{ThinkBLOX} for instruction-driven layout editing in Figure~\ref{fig:edit}. Unlike one-shot methods that often treat an editing request as a new scene generation task, \emph{ThinkBLOX} performs edits in a localized manner. Given an existing layout, the model first identifies the target object or region specified by the instruction, then reasons over the current scene context to produce a new placement action. As a result, it can support common editing operations, including object addition, movement, and replacement, without reconstructing the whole scene.

This localized editing behavior is important for interactive workflows, where users usually expect a small modification to preserve the rest of the layout. As shown in Figure~\ref{fig:edit}, \emph{ThinkBLOX} modifies the intended object while maintaining the surrounding spatial organization. In contrast, existing baselines often trigger global re-layout, change unrelated objects, or even misidentify the target object to be edited. These results demonstrate that the proposed progressive reasoning framework improves not only generation quality but also interactive editability and controllability. More examples of local layout editing and rearrangement are provided in the supplementary material.

\subsection{Ablation Studies}

\begin{figure*}[!t]
  \centering
  \includegraphics[width=0.96\textwidth]{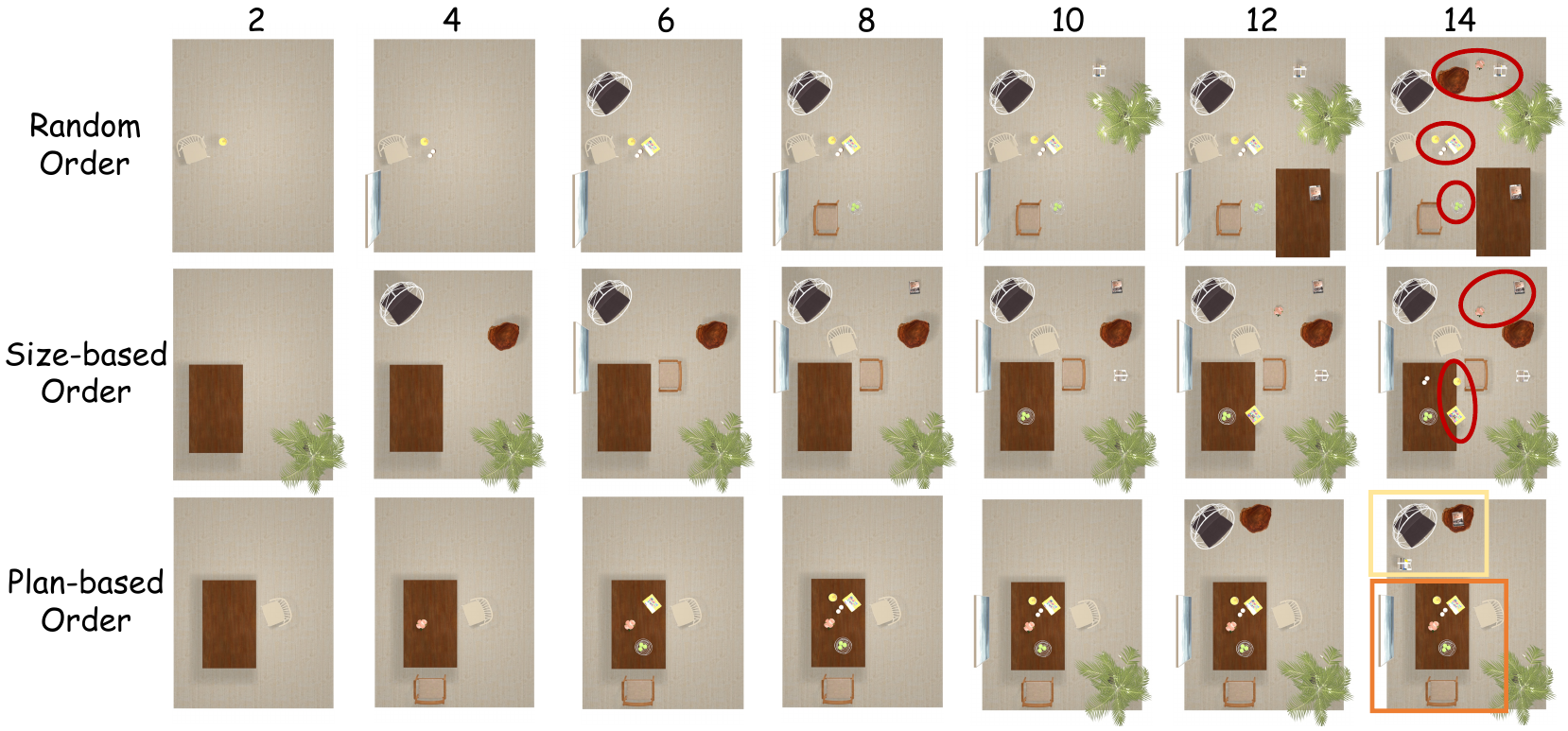}
  \caption{Effect of inference ordering. Plan-based ordering yields clearer functional zones and better organization (yellow/orange), while random and size-based orders lead to cluttered or incorrect placements (red).}
  \label{fig:random-plan}
\end{figure*}

We show ablation studies in Table~\ref{tab:ablation}.
Removing progressive reasoning leads to a clear degradation, showing that decomposing layout planning into sequential reasoning-action steps is important for stable and coherent scene generation. Replacing Tier-Decoupled GDPO with GDPO or GRPO \cite{shao2024deepseekmath} also weakens performance, which validates the benefit of decoupling heterogeneous objectives during policy optimization. In addition, the reward ablations show that each reward set contributes complementary supervision, and removing any one of them harms the overall layout quality. The visual mark ablations further demonstrate that both object-level visual cues and coordinate-level spatial marks are important for grounding object placement. Finally, the inference ordering ablation shows that a plan-based placement order is more effective than heuristic alternatives such as size-based or random ordering. As illustrated in Figure~\ref{fig:random-plan}, plan-based ordering produces more coherent intermediate layouts and better overall scene organization. In general, these results show that the strong performance of \emph{ThinkBLOX} arises from the synergy of progressive reasoning, tier-decoupled optimization, reward decomposition, visual grounding, and global layout plan in the inference stage.
Please refer to the supplementary material for more ablation results.

\begin{table*}[!t]
\centering
\caption{\textbf{Ablation Study.} Our full \emph{ThinkBLOX} model (top row) employs progressive reasoning, Tier-Decoupled GDPO, a complete reward suite, visual representations, and a plan-based inference ordering strategy. Removing progressive reasoning $\mathbf{r}_n$, downgrading the policy optimization algorithms, omitting specific reward signal sets, removing visual marks, or altering the inference ordering strategy consistently degrades both physical plausibility and semantic coherence.}
\label{tab:ablation}
\setlength{\tabcolsep}{2mm} 
\resizebox{\textwidth}{!}{
\begin{tabular}{lccccc}
\toprule
& \multicolumn{2}{c}{Physics} & \multicolumn{2}{c}{Semantics} & Overall Score \\
\cmidrule(lr){2-3} \cmidrule(lr){4-5} \cmidrule(lr){6-6}
& CF$\uparrow$ & IB$\uparrow$ & Pos.$\uparrow$ & Rot.$\uparrow$ & PSA$\uparrow$ \\
\midrule
\textbf{\emph{ThinkBLOX}} & \textbf{81.6 $\pm$ 1.2} & \textbf{91.3 $\pm$ 0.8} & \textbf{79.2 $\pm$ 0.5} & \textbf{72.1 $\pm$ 0.9} & \textbf{54.6 $\pm$ 1.5} \\
\midrule
\multicolumn{6}{l}{\textit{Ablating Progressive Reasoning}} \\
\midrule
SFT (one-shot) w/o $\mathbf{r}_n$ & 39.4 $\pm$ 1.8 & 24.4 $\pm$ 2.0 & 19.3 $\pm$ 1.5 & 29.5 $\pm$ 1.3 & 8.37 $\pm$ 1.8 \\
SFT w/o $\mathbf{r}_n$ & 48.6 $\pm$ 1.5 & 35.7 $\pm$ 1.8 & 23.5 $\pm$ 1.3 & 31.6 $\pm$ 1.3 & 11.7 $\pm$ 1.5 \\
\midrule
\multicolumn{6}{l}{\textit{Ablating Policy Optimization Algorithms}} \\
\midrule
SFT Only (Base) & 63.5 $\pm$ 1.1 & 37.1 $\pm$ 1.8 & 33.6 $\pm$ 1.3 & 56.8 $\pm$ 1.2 & 21.9 $\pm$ 1.5 \\
+ GRPO  & 82.6 $\pm$ 1.8 & 36.2 $\pm$ 1.3 & 23.9 $\pm$ 1.0 & 34.2 $\pm$ 1.1 & 8.65 $\pm$ 2.0 \\
+ GDPO  & 67.8 $\pm$ 1.5 & 68.4 $\pm$ 1.2 & 67.2 $\pm$ 0.8 & 62.5 $\pm$ 0.9 & 47.3 $\pm$ 1.3 \\
\midrule
\multicolumn{6}{l}{\textit{Ablating Reward Sets}} \\
\midrule
w/o $S_{hard}$ & 67.4 $\pm$ 1.8 & 42.8 $\pm$ 1.7 & 57.4 $\pm$ 1.0 & 61.4 $\pm$ 1.3 & 30.3 $\pm$ 2.7 \\
w/o $S_{soft}$ & 77.6 $\pm$ 0.6 & 82.9 $\pm$ 1.9 & 67.4 $\pm$ 0.6 & 54.9 $\pm$ 0.8 & 43.8 $\pm$ 3.0 \\
w/o $S_{cons}$ & 73.5 $\pm$ 1.4 & 79.3 $\pm$ 1.4 & 64.6 $\pm$ 1.0 & 51.8 $\pm$ 1.3 & 36.5 $\pm$ 2.7 \\
\midrule
\multicolumn{6}{l}{\textit{Ablating Visual Marks}} \\
\midrule
w/o Visual Asset      & 62.7 $\pm$ 0.8 & 33.8 $\pm$ 3.9 & 53.2 $\pm$ 0.6 & 47.9 $\pm$ 0.8 & 24.2 $\pm$ 4.8 \\
w/o Visual Coordinate & 54.2 $\pm$ 2.0 & 28.6 $\pm$ 1.7 & 44.8 $\pm$ 1.0 & 23.1 $\pm$ 1.1 & 21.7 $\pm$ 2.7 \\
w/o Any Visual Mark & 49.5 $\pm$ 2.5 & 20.6 $\pm$ 1.8 & 43.7 $\pm$ 1.5 & 18.3 $\pm$ 1.3 & 15.2 $\pm$ 2.8 \\
\midrule
\multicolumn{6}{l}{\textit{Ablating Inference Ordering Strategy}} \\
\midrule
Size-based Order   & 78.8 $\pm$ 0.5 & 85.9 $\pm$ 2.5 & 72.4 $\pm$ 0.5 & 66.4 $\pm$ 0.6 & 47.8 $\pm$ 3.5 \\
Random Order   & 75.6 $\pm$ 0.8 & 78.4 $\pm$ 2.7 & 69.1 $\pm$ 0.9 & 62.2 $\pm$ 1.6 & 41.6 $\pm$ 3.8 \\
\bottomrule
\end{tabular}}
\end{table*}

\subsection{Human Evaluation}

\begin{figure}[!t]
  \centering
  \includegraphics[width=\columnwidth]{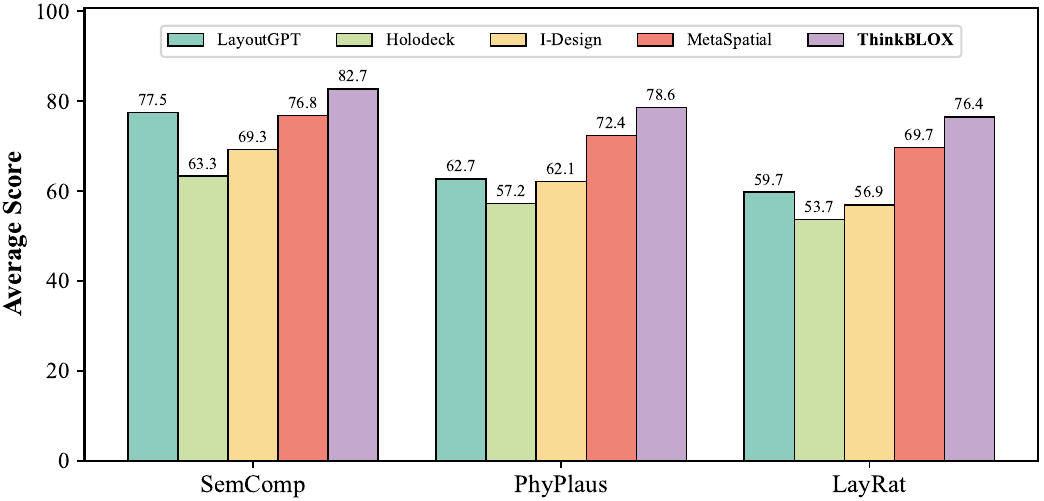}
  \caption{Average human evaluation results across 12 room categories. \emph{ThinkBLOX} achieves the highest average scores under all three metrics: semantic compliance (SemComp), physical plausibility (PhyPlaus), and layout rationality (LayRat).}
  \label{fig:all_metrics_bar}
\end{figure}
We further conduct a user study to assess the perceptual quality of the generated layouts. Specifically, we design an online questionnaire and collect 33 valid responses. The evaluation covers 12 representative scenes, each associated with 5 layout visualizations produced by different methods, resulting in a total of 60 evaluated images. To ensure fairness, all results are anonymized and presented in random order. Participants evaluate each result under three criteria: Semantic Compliance, reflecting the consistency between the generated layout and the input instruction; Physical Plausibility, evaluating whether the placements are physically feasible, e.g., properly supported and free of noticeable interpenetration; and Layout Rationality, measuring whether the arrangement of objects is functionally and spatially reasonable.

As shown in Figure~\ref{fig:all_metrics_bar}, our method consistently receives the best human ratings across all three dimensions. More specifically, this advantage remains stable across all 12 room categories, suggesting that the proposed framework generalizes robustly to diverse indoor scenes rather than benefiting only from a small subset of easy cases. Detailed per-room human evaluation results are provided in the supplementary material. The gains in \textit{Physical Plausibility} and \textit{Layout Rationality} indicate that our progressive reasoning and placement strategy helps avoid common one-shot failure cases, such as collisions, unsupported objects, and functionally awkward arrangements. Meanwhile, the improvement in \textit{Semantic Compliance} further shows that the generated layouts align more faithfully with user instructions and scene intent. Detailed evaluation criteria and per-room human evaluation scores are provided in the supplementary material.

\section{Conclusion}

In this paper, we presented \emph{ThinkBLOX}, a VLM-based progressive reason-then-act framework for interactive 3D indoor scene generation and editing. By formulating layout synthesis as a stepwise placement process, \emph{ThinkBLOX} supports both coherent scene generation and flexible object-level editing without requiring full scene reconstruction. To enable this capability, we constructed ThinkBLOX-Data-200K for progressive reasoning and introduced Tier-Decoupled GDPO for stable multi-objective optimization over physical validity, semantic plausibility, and reasoning--action consistency. Extensive experiments show that \emph{ThinkBLOX} achieves strong physical plausibility, semantic coherence, and interactive editability.

Our method still has several limitations. First, progressive inference is about three to five times slower than one-shot generation. Second, the reward design follows general principles of interior design that may not fully capture subtle or personalized preferences. In future work, we plan to improve inference efficiency and explore stronger reward learning with human feedback.

\bibliographystyle{IEEEtran}
\bibliography{main}



\section*{Supplementary material}

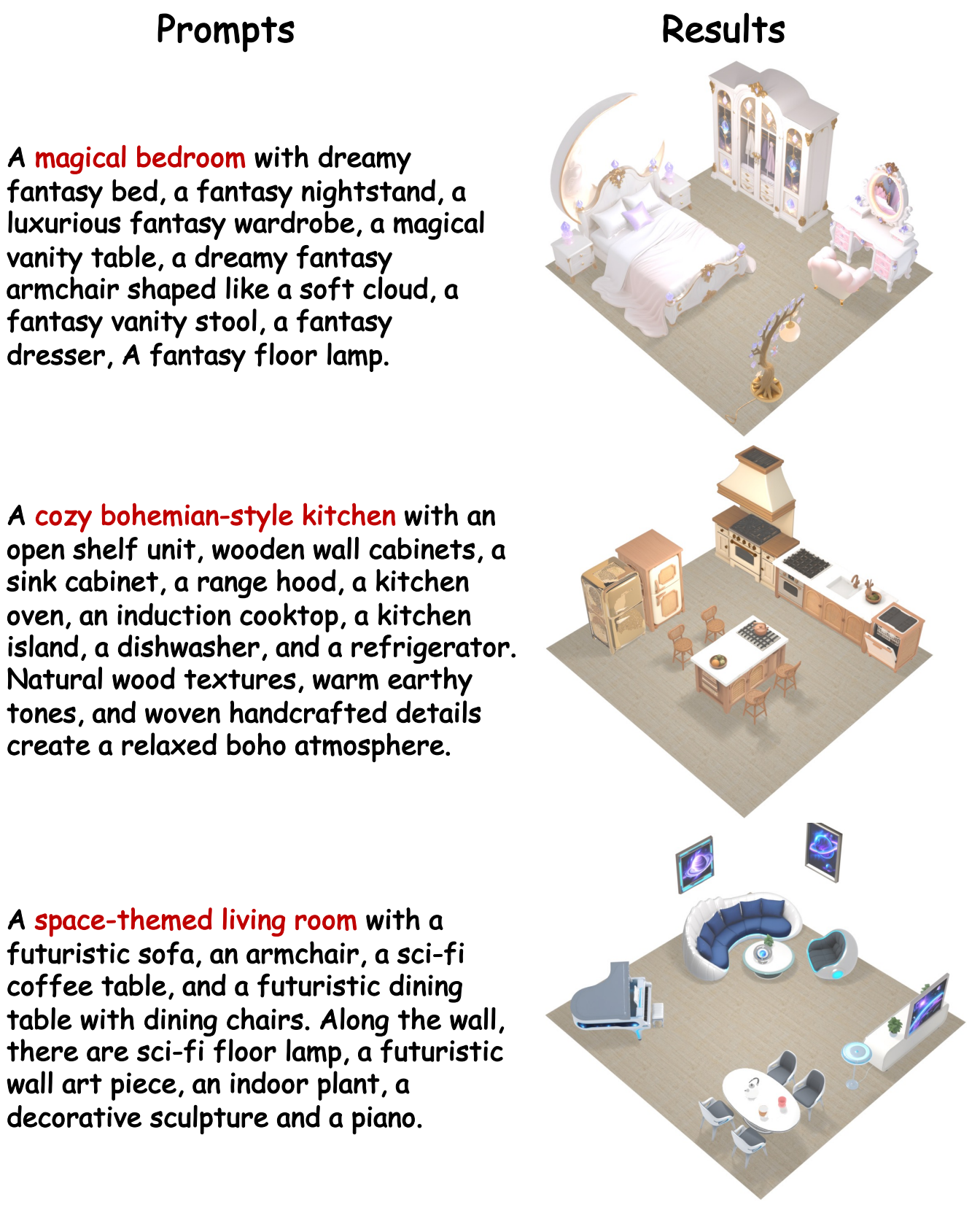


Fig. 5. Additional examples of global layout generation from text prompts. *ThinkBLOX* generates complete indoor layouts for semantically and stylistically diverse scenes, including a magical bedroom, a bohemian-style kitchen, and a space-themed living room. The results illustrate the ability of our progressive reasoning framework to translate user intent into coherent scene organization and style-consistent object placement.

no obvious deviation or omission.

- 80-61: Good semantic compliance The scene generally matches the input description. Most key objects and semantic requirements are correctly expressed, although there may still be a small number of inaccurate, missing, or imperfect details.
- 60-41: Moderate semantic compliance The scene only partially matches the input description. Some core semantics are reflected, but there are noticeable problems in object selection, scene functionality, style expression, or detailed presentation.
- 40-21: Poor semantic compliance The scene has a relatively low degree of consistency with the input description. Many key semantic elements are not correctly reflected, and there are clear deviations in room purpose, object configuration, or overall style.
- 20-1: Very poor semantic compliance The scene largely fails to reflect the input description. Most key objects, functional requirements,

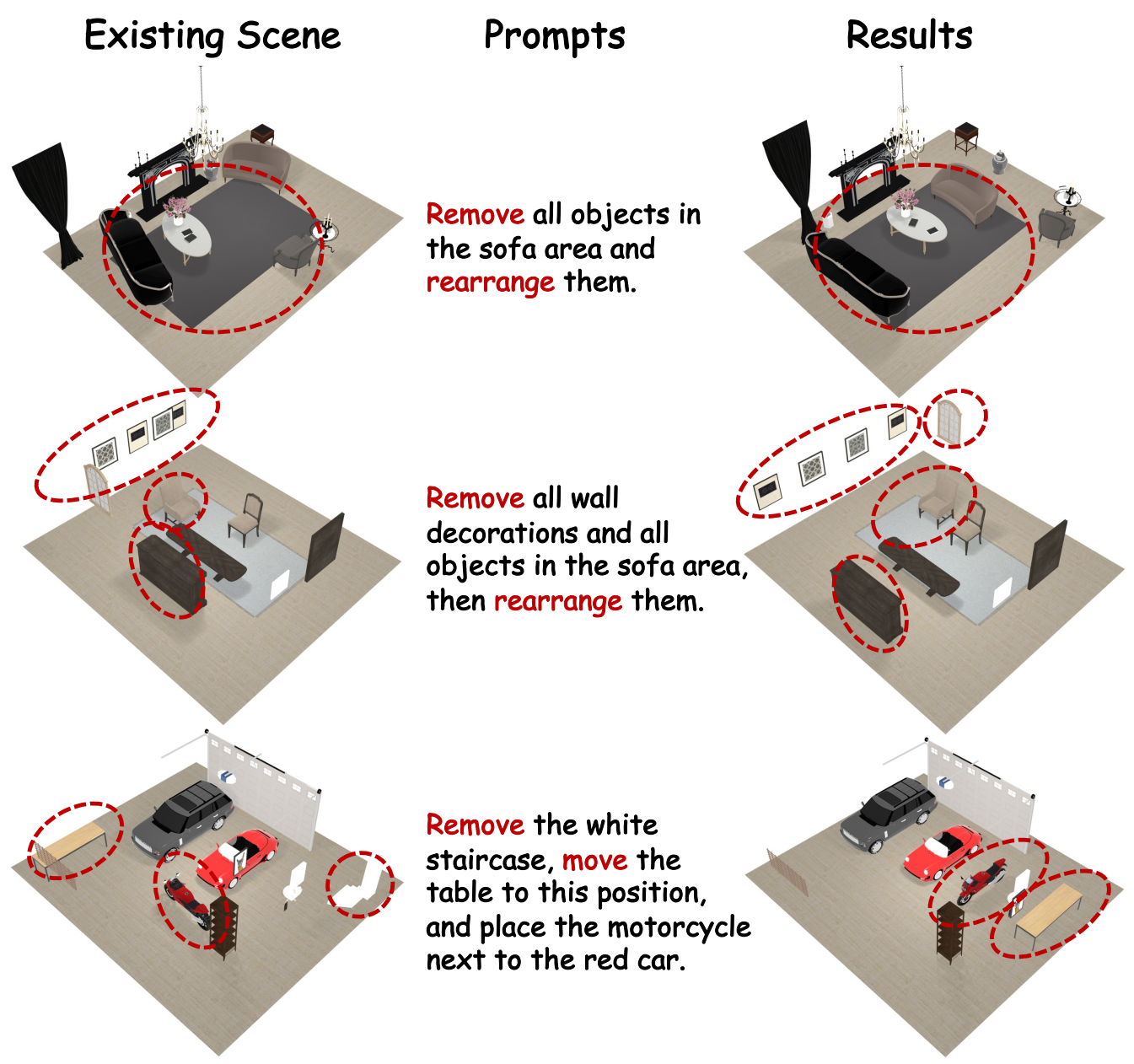


Fig. 6. Additional qualitative examples of local layout rearrangement under compound editing instructions. *ThinkBLOX* supports localized modifications such as object removal, region-level rearrangement, and object relocation, while keeping the remaining scene stable. The examples demonstrate that the proposed progressive reasoning framework can flexibly handle multi-step editing commands on existing scenes.

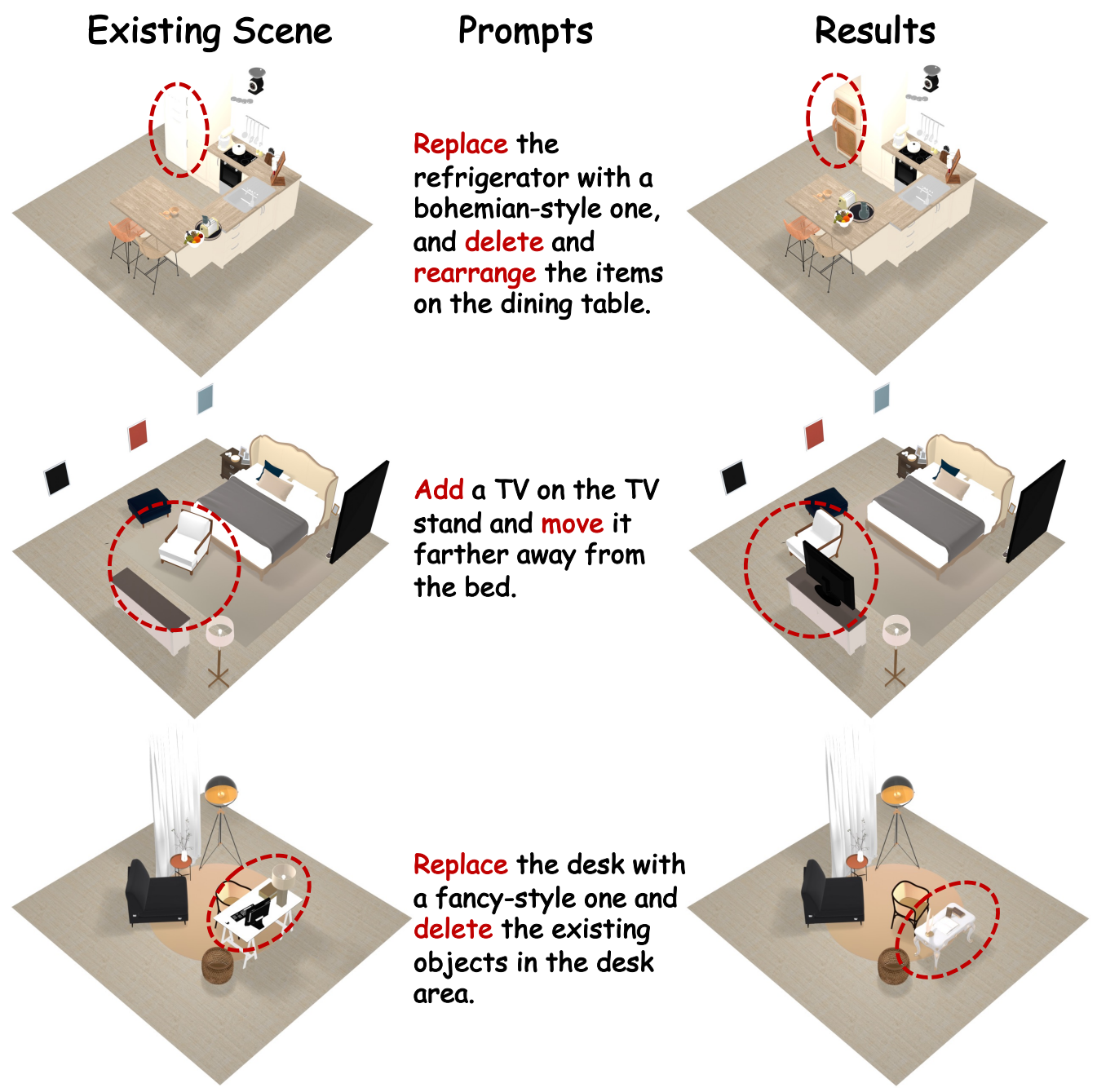


Fig. 7. More examples of fine-grained local layout editing. *ThinkBLOX* can perform object replacement, object addition, and local reorganization according to user instructions, without re-generating the whole scene. The examples show localized edits in kitchen, bedroom, and study scenarios, further illustrating the flexibility of our progressive reasoning pipeline for interactive scene modification.

and semantic intentions are missing or incorrect, and the overall content is seriously inconsistent

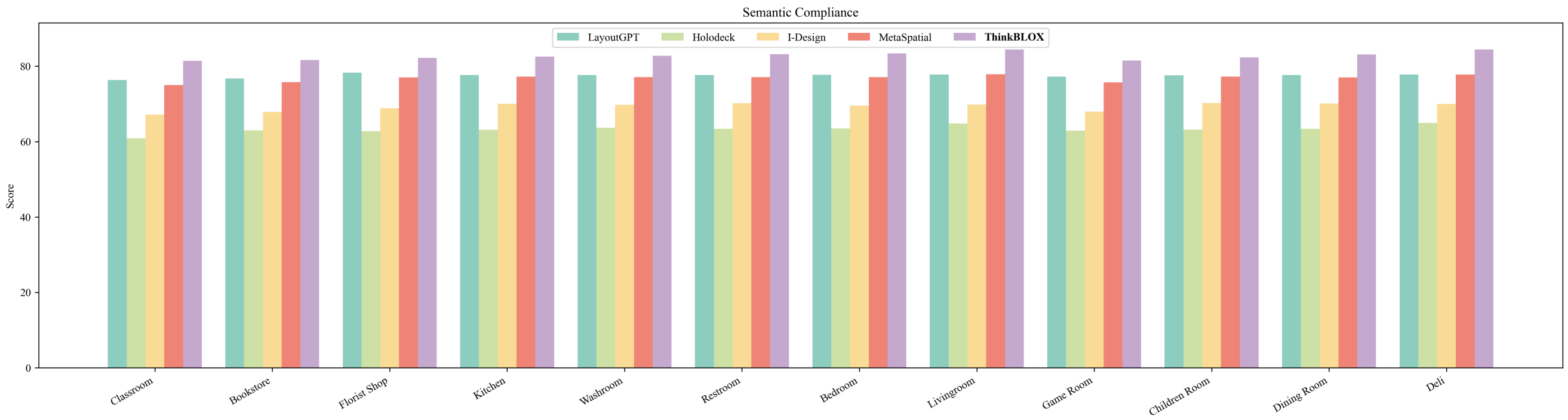


(a) Semantic Compliance

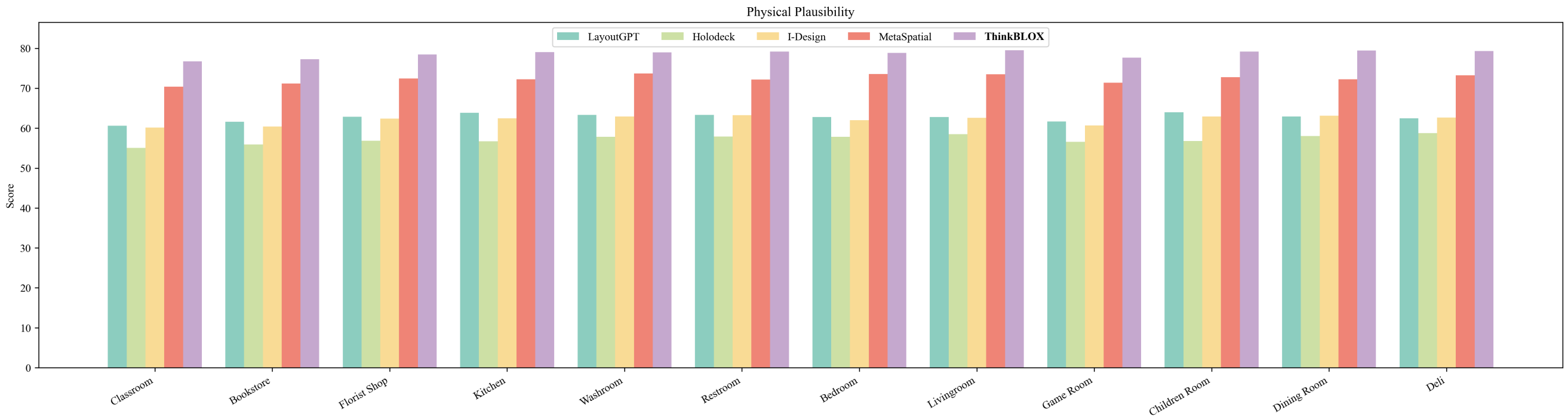


(b) Physical Plausibility

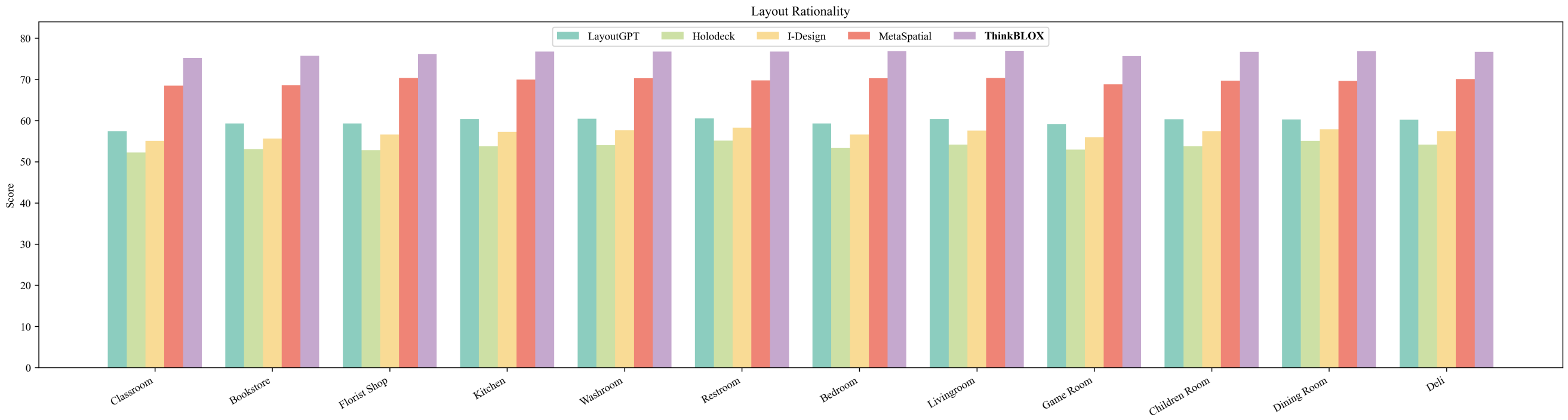


(c) Layout Rationality

Fig. 8. Comparison of five methods across 12 room categories under three evaluation metrics.

with the target description.

2) **Physical Plausibility**
This criterion evaluates whether object placements in the scene satisfy basic physical commonsense and spatial constraints, such as the absence of severe interpenetration, floating objects, out-of-bound placements, unstable support, or obviously infeasible configurations.

- 100–81: Very high physical plausibility Object placements fully comply with basic physical principles. There are almost no issues such as collisions, floating, interpenetration, boundary violations, or unreasonable support, making the overall scene realistic and credible.
- 80–61: Good physical plausibility Most object placements are physically reasonable. Only a small number of minor issues may exist, but they do not significantly affect the realism or overall acceptability of the scene.
- 60–41: Moderate physical plausibility The scene contains a certain number of physical problems, such as local collisions, slight floating, unstable placements, or improper boundary handling. These issues somewhat affect the realism of the scene.

- 40–21: Poor physical plausibility The scene contains many obvious physical errors, such as large object intersections, clear floating, severe boundary violations, or incorrect support relationships, which significantly reduce overall feasibility.
- 20–1: Very poor physical plausibility The scene seriously violates physical commonsense. A large number of objects exhibit collisions, interpenetration, floating, or impossible placements, making the scene almost entirely unrealistic and infeasible.

3) **Layout Rationality**
This criterion evaluates whether the spatial organization and object arrangement in the scene are reasonable, including whether functional zones are clearly defined, whether related objects are properly placed, whether the layout is convenient for use, and whether the overall space appears coordinated and natural.
   - 100–81: Very high layout rationality The scene layout is highly reasonable. Functional zones are clearly organized, the relative positions of related objects are natural and coherent, space is used efficiently, and the layout follows realistic interior design logic.
   - 80–61: Good layout rationality The overall scene layout is fairly reasonable. Most object placements are consistent with common usage habits and design principles, with only a few local regions being less ideal.
   - 60–41: Moderate layout rationality The scene exhibits basic layout logic, but there are several suboptimal arrangements, such as unclear functional zoning, inconvenient local accessibility, or unnatural relationships among objects.
   - 40–21: Poor layout rationality The scene layout has obvious problems. Multiple objects are placed in ways that do not follow normal usage logic, and the spatial organization is confusing, harming functionality, accessibility, or overall coherence.
   - 20–1: Very poor layout rationality The scene has almost no reasonable layout organization. Objects are arranged chaotically, lacking basic functional planning and spatial logic, which severely affects both usability and visual coherence.



\end{document}